\documentclass[11pt,a4paper]{article}
\usepackage[hyperref]{acl2018}
\usepackage{times}
\usepackage{latexsym}

\usepackage{graphicx}
\usepackage{wrapfig}
\usepackage{url}
\usepackage{amsmath}
\usepackage{amssymb}
\usepackage{color,xcolor,colortbl}
\usepackage{enumitem}
\usepackage{algorithm}
\usepackage{algorithmic}
\usepackage[font={small}]{caption}
\usepackage{bm}
\usepackage{booktabs}

\DeclareMathOperator*{\argmax}{arg\,max\,}
\newcommand\numberthis{\addtocounter{equation}{1}\tag{\theequation}}

\definecolor{darkblue}{rgb}{0.0, 0.0, 0.55}
\newenvironment{fontppl}{\fontfamily{ppl}\selectfont}{\par} 

\aclfinalcopy 
\def\aclpaperid{46} 

\title{Reinforced Extractive Summarization with Question-Focused Rewards}

\author{Kristjan Arumae \quad Fei Liu\\ 
  Department of Computer Science\\
  University of Central Florida, 
  Orlando, FL 32816, USA\\
  {\tt k.arumae@knights.ucf.edu \quad feiliu@cs.ucf.edu}}

\date{}

\begin{document}

\maketitle

\begin{abstract}

We investigate a new training paradigm for extractive summarization.
Traditionally, human abstracts are used to derive goldstandard labels for extraction units.
However, the labels are often inaccurate, because human abstracts and source documents cannot be easily aligned at the word level.
In this paper we convert human abstracts to a set of Cloze-style comprehension questions.
System summaries are encouraged to preserve salient source content useful for answering questions and share common words with the abstracts.
We use reinforcement learning to explore the space of possible extractive summaries and introduce a question-focused reward function to promote concise, fluent, and informative summaries.
Our experiments show that the proposed method is effective. 
It surpasses state-of-the-art systems on the standard summarization dataset.

\end{abstract}

\section{Introduction}
\label{sec:introduction}

We study extractive summarization in this work where salient word sequences are extracted from the source document and concatenated to form a summary~\cite{Nenkova:2011}.
Existing supervised approaches to extractive summarization frequently use human abstracts to create annotations for extraction units~\cite{Gillick:2009:NAACL,Li:2013:EMNLP,Cheng:2016}. 
E.g., a source word is labelled 1 if it appears in the abstract, 0 otherwise.
Despite the usefulness, there are two issues with this scheme.
First, a vast majority of the source words are tagged 0s, only a small portion are 1s.
This is due to the fact that human abstracts are short and concise; they often contain words not present in the source.
Second, not all labels are accurate.
Source words that are labelled 0 may be paraphrases, generalizations, or otherwise related to words in the abstracts.
These source words are often mislabelled.
Consequently, leveraging human abstracts to provide supervision for extractive summarization remains a challenge.

Neural abstractive summarization can alleviate this issue by allowing the system to either copy words from the source texts or generate new words from a vocabulary~\cite{Rush:2015,Nallapati:2016,See:2017}. 
While the techniques are promising, they face other challenges, such as ensuring the summaries remain faithful to the original. 
Failing to reproduce factual details has been revealed as one of the main obstacles for neural abstractive summarization~\cite{Cao:2018,Song:2018}.
This study thus chooses to focus on neural extractive summarization.

\begin{table}
\centering
\setlength{\tabcolsep}{5pt}
\renewcommand{\arraystretch}{1.1}
\begin{scriptsize}
\begin{fontppl}
\begin{tabular}{|p{2.8in}|}
\hline
\textbf{Source Document}\\[0.7mm]
\emph{\textbf{The first doses of the Ebola vaccine were on a commercial flight to West Africa and were expected to arrive on Friday}}, according to a spokesperson from GlaxoSmithKline (GSK) one of the companies that has created the vaccine with the National Institutes of Health.\\[0.7mm]

Another vaccine from Merck and NewLink will also be tested.\\[0.7mm]

``Shipping the vaccine today is a major achievement and shows that we remain on track with the accelerated development of our candidate Ebola vaccine,'' Dr. Moncef Slaoui, chairman of global vaccines at GSK said in a company release. (Rest omitted.)\\[0.7mm]
\hline
\hline
\textbf{Abstract}\\[0.7mm]
The first vials of an Ebola vaccine should land in Liberia Friday\\
\hline
\hline
\textbf{Questions}\\[0.7mm]
\textbf{Q:} The first vials of an \_\_\_\_ vaccine should land in Liberia Friday\\
\textbf{Q:} The first vials of an Ebola vaccine should \_\_\_\_ in Liberia Friday\\
\textbf{Q:} The first vials of an Ebola vaccine should land in \_\_\_\_ Friday\\
\hline
\end{tabular}
\end{fontppl}
\end{scriptsize}
\caption{Example source document, the top sentence of the abstract, and system-generated Cloze-style questions. Source content related to the abstract is \textit{italicized}.}
\label{tab:example}
\vspace{-0.1in}
\end{table}

We explore a new training paradigm for extractive summarization.
We convert human abstracts to a set of Cloze-style comprehension questions, where the question body is a sentence of the abstract with a blank, and the answer is an entity or a keyword.
Table~\ref{tab:example} shows an example.
Because the questions cannot be answered by applying general world knowledge, system summaries are encouraged to preserve salient source content that is relevant to the questions ($\approx$ human abstract) such that the summaries can work as a document surrogate to predict correct answers.
We use an attention mechanism to locate segments of a summary that are relevant to a given question so that the summary can be used to answer multiple questions.

This study extends the work of~\cite{Lei:2016} to use reinforcement learning to explore the space of extractive summaries. 
While the original work focuses on generating rationales to support supervised classification, 
the goal of our study is to produce fluent, generic document summaries. 
The question-answering (QA) task is designed to fulfill this goal and the QA performance is only secondary. 
Our research contributions can be summarized as follows:
\begin{itemize}[topsep=3pt,itemsep=-1pt,leftmargin=*]

\item we investigate an alternative training scheme for extractive summarization where the summaries are encouraged to be semantically close to human abstracts in addition to sharing common words;

\item we compare two methods to convert human abstracts to Cloze-style questions and investigate its impact on QA and summarization performance. Our results surpass those of previous systems on a standard summarization dataset.

\end{itemize}

\section{Related Work}

This study focuses on generic summarization.
It is different from the query-based summarization~\cite{Daume:2006:ACL,Dang:2008}, where systems are trained to select text pieces related to \emph{predefined} queries. 
In this work we have no predefined queries but the system carefully generates questions from human abstracts and learns to produce generic summaries that are capable of answering all questions.

Cloze questions have been used in reading comprehension~\cite{Richardson:2013,Weston:2016,Mostafazadeh:2016,Rajpurkar:2016} to test the system's ability to perform reasoning and language understanding. 
Hermann et al.~\shortcite{Hermann:2015} describe an approach to extract (context, question, answer) triples from news articles.
Our work draws on this approach to automatically create questions from human abstracts.

Reinforcement learning (RL) has been recently applied to a number of NLP applications, including dialog generation~\cite{Li:2017:GAN}, machine translation (MT)~\cite{Ranzato:2016,Gu:2018}, question answering~\cite{Choi:2017}, and summarization and sentence simplification~\cite{Zhang:2017,Paulus:2017,Chen:2018:ACL,Narayan:2018}.
This study leverages RL to explore the space of possible extractive summaries. 
The summaries are encouraged to preserve salient source content useful for answering questions as well as sharing common words with the abstracts.

\begin{figure*}
\centering
\includegraphics[width=5.9in]{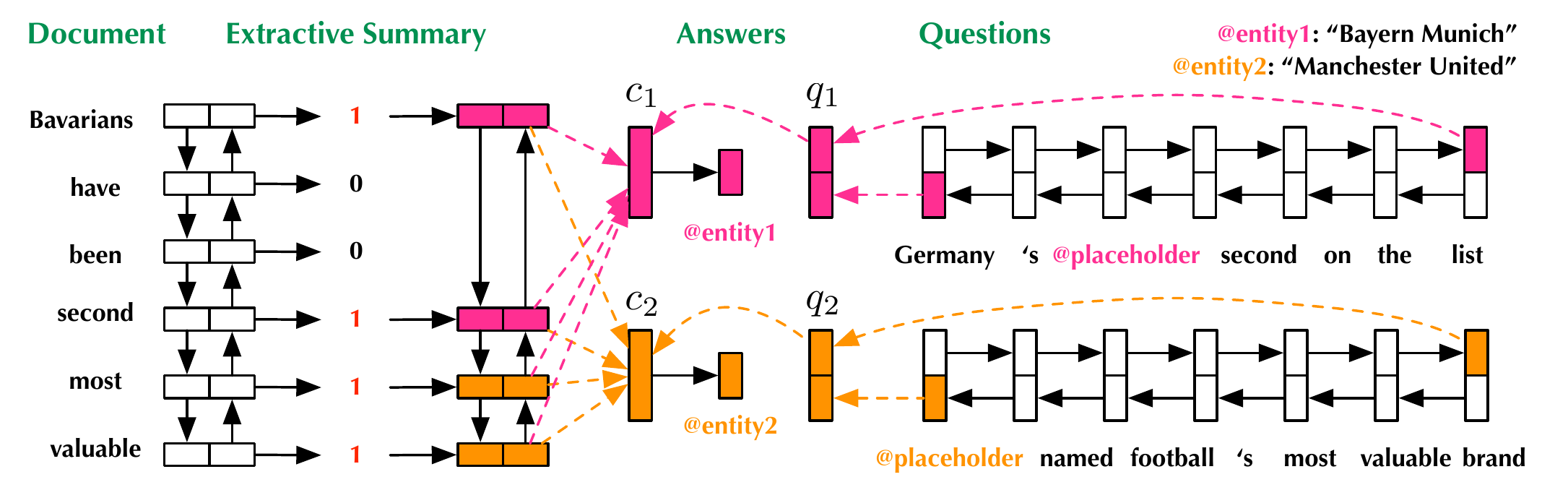}
\caption{System framework. The model uses an extractive summary as a document surrogate to answer important questions about the document. The questions are automatically derived from the human abstract.}
\label{fig:architecture}
\vspace{-0.1in}
\end{figure*}

\section{Our Approach}
\label{sec:approach}

Given a source document $X$, our system generates a summary $Y=(y_1, y_2, \cdots, y_{|Y|})$ by identifying consecutive sequences of words: $y_t$ is 1 if the $t$-th source word is included in the summary, 0 otherwise.
In this section we investigate a question-oriented reward $\mathcal{R}(Y)$ that encourages summaries to contain sufficient content useful for answering key questions about the document  (\S\ref{sec:rewards}); we then use reinforcement learning to explore the space of possible extractive summaries (\S\ref{sec:sampling}).

\subsection{Question-Focused Reward}
\label{sec:rewards}

We reward a summary if it can be used as a document surrogate to answer important questions.
Let $\{(Q_k,e_k^*)\}_{k=1}^K$ be a set of question-answer pairs for a source document, where $e_k^*$ is the ground-truth answer corresponding to an entity or a keyword.
We encode the question $Q_k$ into a vector: $\mathbf{q}_k = \mbox{Bi-LSTM}(Q_k) \in \mathbb{R}^d$ using a bidirectional LSTM, where the last outputs of the forward and backward passes are concatenated to form a question vector.
We use the same Bi-LSTM to encode the summary $Y$ to a sequence of vectors: $(\mathbf{h}_1^S, \mathbf{h}_2^S, \cdots, \mathbf{h}_{|S|}^S) = \mbox{Bi-LSTM}(Y)$, where $|S|$ is the number of words in the summary; 
$\mathbf{h}_t^S \in \mathbb{R}^d$ is the concatenation of forward and backward hidden states at time step $t$.
Figure~\ref{fig:architecture} provides an illustration of the system framework.

An attention mechanism is used to locate parts of the summary that are relevant to $Q_k$.
We define $\alpha_{k,i} \propto \exp(\mathbf{q}_k \mathbf{W}^a \mathbf{h}_i^S)$ to represent the importance of the $i$-th summary word ($\mathbf{h}_i^S$) to answering the $k$-th question ($\mathbf{q}_k$), characterized by a bilinear term~\cite{Chen:2016:CNN}.
A context vector $\mathbf{c}_k$ is constructed as a weighted sum of all summary words relevant to the $k$-th question, and it is used to predict the answer. 
We define the QA reward $\mathcal{R}_a(Y)$ as the log-likelihood of correctly predicting all answers.
$\{\mathbf{W}^a, \mathbf{W}^c\}$ are learnable model parameters.
\begin{align*}
& \alpha_{k,i} = \frac{\exp(\mathbf{q}_k \mathbf{W}^a \mathbf{h}_i^S)}{\sum_{i=1}^{|S|} \exp(\mathbf{q}_k \mathbf{W}^a \mathbf{h}_i^S)}
\numberthis\label{eq:alpha_k_i}\\
& \mathbf{c}_k = \sum_{i=1}^{|S|} \alpha_{k,i}\mathbf{h}_i^S
\numberthis\label{eq:c_k}\\
& P(e_k|Y,Q_k)= \mbox{softmax}(\mathbf{W}^c \mathbf{c}_k)
\numberthis\label{eq:p_e}\\
& \mathcal{R}_a(Y) = \frac{1}{K}\sum_{k=1}^K \log P(e_k^*|Y, Q_k)
\numberthis\label{eq:r_a}
\end{align*}

In the following we describe approaches to obtain a set of question-answer pairs $\{(Q_k,e_k^*)\}_{k=1}^K$ from a human abstract.
In fact, this formulation has the potential to make use of multiple human abstracts (subject to availability) in a unified framework; in that case, the QA pairs will be extracted from all abstracts. 
According to Eq.~(\ref{eq:r_a}), the system is optimized to generate summaries that preserve salient source content sufficient to answer \emph{all questions} ($\approx$ human abstract).

We expect to harvest one question-answer pair from each sentence of the abstract. More are possible, but the QA pairs will contain duplicate content. 
There are a few other noteworthy issues.
If we do not collect any QA pairs from a sentence of the abstract, its content will be left out of the system summary. It is thus crucial for the system to extract at least one QA pair from \emph{any} sentence in an automatic manner.
Further, the questions must not be answered by simply applying general world knowledge.
We expect the adequacy of the summary to have a direct influence on whether or not the questions will be correctly answered.
Motivated by these considerations, we perform the following steps. 
We split a human abstract to a set of sentences, identify an answer token from each sentence, then convert the sentence to a question by replacing the token with a placeholder, yielding a Cloze question.
We explore two approaches to extract answer tokens:
\begin{itemize}[topsep=3pt,itemsep=0pt]
\item \textit{Entities.} We extract four types of named entities \{\textsc{per}, \textsc{loc}, \textsc{org}, \textsc{misc}\} from sentences and treat them as possible answer tokens. 
\item \textit{Keywords.} This approach identifies the \textsc{root} word of a sentence dependency parse tree and treats it as a keyword-based answer token. Not all sentences contain entities, but every sentence has a root word; it is often the main verb of the sentence.
\end{itemize}
We obtain $K$ question-answer pairs from each human abstract, one pair per sentence. If there are less than $K$ sentences in the abstract, the QA pairs of the top sentences will be duplicated, with the assumption that the top sentences are more important than others.
If multiple entities reside in a sentence, we randomly pick one as the answer token; otherwise if there are no entities, we use the root word instead.

To ensure that the extractive summaries are concise, fluent, and close to the original wording, we add additional components to the reward function:
(i) we define $\mathcal{R}_s(Y) = |\frac{1}{|Y|}\sum_{t=1}^{|Y|}y_t - \delta|$ to restrict the summary size.
We require the percentage of selected source words to be close to a predefined threshold $\delta$. 
This constraint works well at restricting length, with the average summary size adhering to this percentage;
(ii) we further introduce $\mathcal{R}_f(Y) = \sum_{t=2}^{|Y|} |y_t - y_{t-1}|$ to encourage the summaries to be fluent.
This component is adopted from~\cite{Lei:2016}, where few 0/1 switches between $y_{t-1}$ and $y_t$ indicates the system is selecting consecutive word sequences;
(iii) we encourage system and reference summaries to share common bigrams. This practice has shown success in earlier studies~\cite{Gillick:2009:NAACL}.
$\mathcal{R}_b(Y)$ is defined as the percentage of reference bigrams successfully covered by the system summary.
These three components together ensure the well-formedness of extractive summaries.
The final reward function $\mathcal{R}(Y)$ is a linear interpolation of all the components; $\gamma,\alpha,\beta$ are coefficients and we describe their parameter tuning in \S\ref{sec:experiments}.
{\medmuskip=1mu
\thinmuskip=1mu
\thickmuskip=1mu
\nulldelimiterspace=0pt
\scriptspace=0pt
\begin{align*}
&\mathcal{R}(Y) = \mathcal{R}_a(Y) + \gamma\mathcal{R}_b(Y) - \alpha\mathcal{R}_f(Y) - \beta\mathcal{R}_s(Y)
\numberthis\label{equ:r_y}
\end{align*}}

\vspace{-0.25in}
\subsection{Reinforcement Learning}
\label{sec:sampling}

In the following we seek to optimize a policy $P(Y|X)$ for generating extractive summaries so that the expected reward $\mathbb{E}_{P(Y|X)}[\mathcal{R}(Y)]$ is maximized.
Taking derivatives of this objective with respect to model parameters $\theta$ involves repeatedly sampling summaries $\hat{Y} = (\hat{y}_1, \hat{y}_2,\cdots,\hat{y}_{|Y|})$ (illustrated in Eq.~(\ref{eq:nabla_r})).
In this way reinforcement learning exploits the space of extractive summaries of a source document.
\begin{align*}
& \nabla_\theta \mathbb{E}_{P(Y|X)}[\mathcal{R}(Y)]\\
&= \mathbb{E}_{P(Y|X)}[\mathcal{R}(Y)\nabla_\theta\log P(Y|X)]\\
&\approx \textstyle\frac{1}{N}\sum_{n=1}^N \mathcal{R}(\hat{Y}^{(n)}) \nabla_\theta \log P(\hat{Y}^{(n)}|X)
\numberthis\label{eq:nabla_r}
\end{align*}

To calculate $P(Y|X)$ and then sample $\hat{Y}$ from it, we use a bidirectional LSTM to encode a source document to a sequence of vectors: $(\mathbf{h}_1^D, \mathbf{h}_2^D, \cdots, \mathbf{h}_{|X|}^D) = \mbox{Bi-LSTM}(X)$.
Whether to include the $t$-th source word in the summary ($\hat{y}_t$) thus can be decided based on $\mathbf{h}_t^D$.
However, we also want to accommodate the previous $t$-1 sampling decisions ($\hat{y}_{1:t-1}$) to improve the fluency of the extractive summary.
Following~\cite{Lei:2016}, we introduce a single-direction LSTM encoder whose hidden state $\mathbf{s}_t$ tracks the sampling decisions up to time step $t$ (Eq.~\ref{eq:s_t}). 
It represents the semantic meaning encoded in the current summary.
To sample the $t$-th word, we concatenate the two vectors $[\mathbf{h}_t^D||\mathbf{s}_{t-1}]$ and use it as input to a feedforward layer with sigmoid activation to estimate $\hat{y}_t \sim P({y}_t|\hat{y}_{1:t-1}, X)$ (Eq.~\ref{eq:y_t}).
\begin{align*}
& P({y}_t|\hat{y}_{1:t-1}, X) = \sigma(\mathbf{W}^h[\mathbf{h}_t^D||\mathbf{s}_{t-1}] + b^h)
\numberthis\label{eq:y_t}\\
& \mathbf{s}_t = \mbox{LSTM}([\mathbf{h}_t^D||\hat{y}_t], \mathbf{s}_{t-1})
\numberthis\label{eq:s_t}\\
& P(\hat{Y}|X) = \textstyle\prod_{t=1}^{|Y|} P(\hat{y}_t|\hat{y}_{1:t-1}, X)
\numberthis\label{eq:p_y}
\end{align*}
Note that Eq.~(\ref{eq:y_t}) can be pretrained using goldstandard summary sequence $Y^* = ({y}_1^*, {y}_2^*, \cdots, {y}_{|Y|}^*)$ to minimize the word-level cross-entropy loss, where we set ${y}_t^*$ as 1 if (${x}_t$, ${x}_{t+1}$) is a bigram in the human abstract. 
For reinforcement learning, our goal is to optimize the policy $P(Y|X)$ using the reward function $\mathcal{R}(Y)$ (\S\ref{sec:rewards}) during the training process. Once the policy $P(Y|X)$ is learned, we do not need the reward function (or any QA pairs) at test time to generate generic summaries.
Instead we choose $\hat{y}_t$ that yields the highest probability $\hat{y}_t=\argmax P({y}_t|\hat{y}_{1:t-1}, X)$.

\begin{table}[t]
\setlength{\tabcolsep}{5pt}
\renewcommand{\arraystretch}{1.1}
\centering
\begin{small}
\begin{tabular}{|l|rrr|}
\hline
\textbf{System} & \textbf{R-1} & \textbf{R-2} & \textbf{R-L}\\
\hline
\hline
LSA{\scriptsize~\cite{Steinberger:2004}} & 21.2 & 6.2 & 14.0\\
LexRank{\scriptsize~\cite{Erkan:2004}} & 26.1 & 9.6 & 17.7\\
TextRank{\scriptsize~\cite{Mihalcea:2004}} & 23.3 & 7.7 & 15.8\\
SumBasic{\scriptsize~\cite{Vanderwende:2007}} & 22.9 & 5.5 & 14.8\\
KL-Sum{\scriptsize~\cite{Haghighi:2009}} & 20.7 & 5.9 & 13.7\\
Distraction-M3{\scriptsize~\cite{Chen:2016}} & 27.1 & 8.2 & 18.7\\
Seq2Seq w/ Attn{\scriptsize~\cite{See:2017}} & 25.0 & 7.7 & 18.8\\
Pointer-Gen w/ Cov{\scriptsize~\cite{See:2017}} & 29.9 & 10.9 & 21.1\\
Graph-based Attn{\scriptsize~\cite{Tan:2017}} & 30.3 & 9.8 & 20.0\\
\hline
\hline
Extr+EntityQ (this paper) & \textbf{31.4} & \textbf{11.5} & \textbf{21.7}\\
Extr+KeywordQ (this paper) & \textbf{31.7} & \textbf{11.6} & \textbf{21.5}\\
\hline
\end{tabular}
\end{small}
\caption{Results on the CNN test set (full-length F1 scores).}
\label{tab:results_test}
\vspace{-0.1in}
\end{table}

\section{Experiments}
\label{sec:experiments}

All training, validation, and testing was performed using the CNN dataset~\cite{Hermann:2015, Nallapati:2016} containing news articles paired with human-written highlights (i.e., abstracts).
We observe that a source article contains 29.8 sentences and an abstract contains 3.54 sentences on average.
The train/valid/test splits contain 90,266, 1,220, 1,093 articles respectively.

\subsection{Hyperparameters}
The hyperparameters, tuned on the validation set, include the following:
the hidden state size of the Bi-LSTM is 256; 
the hidden state size of the single-direction LSTM encoder is 30.
Dropout rate \cite{Dropout}, used twice in the sampling component, is set to 0.2.
The minibatch size is set to 256.
We apply early stopping on the validation set, where the maximum number of epochs is set to 50.
Our source vocabulary contains 150K words; words not in the vocabulary are replaced by the {\fontfamily{fontppl}\selectfont $\langle\mbox{unk}\rangle$} token.
We use 100-dimensional word embeddings, initialized by GloVe~\cite{Pennington:2014} and remain trainable.
We set $\beta$ = 2$\alpha$ and select the best $\alpha \in \{\underline{10}, 20, 50\}$ and $\gamma \in \{5, 6, 7, \underline{8}\}$ using the valid set (best value underlined).
The maximum length of input is set to 100 words; $\delta$ is set to be 0.4 ($\approx$40 words).
We use the Adam optimizer~\cite{Kingma:2015} with an initial learning rate of 1e-4 and halve the learning rate if the objective worsens beyond a threshold ($> 10\%$).  As mentioned we utilized a bigram based pretraining method.  We found that this stabilized the training of the full model.

\subsection{Results}
We compare our methods with state-of-the-art published systems, including both extractive and abstractive approaches (their details are summarized below).
We experiment with two variants of our approach. ``EntityQ'' uses QA pairs whose answers are named entities. ``KeywordQ'' uses pairs whose answers are sentence root words.
According to the R-1, R-2, and R-L scores~\cite{Lin:2004} presented in Table~\ref{tab:results_test}, both methods are superior to the baseline systems on the benchmark dataset, yielding 11.5 and 11.6 R-2 F-scores, respectively.

\begin{itemize}[topsep=6pt,itemsep=-1pt,leftmargin=*]
\item \textbf{LSA}~\cite{Steinberger:2004} uses the latent semantic analysis technique to identify semantically important sentences.

\item \textbf{LexRank}~\cite{Erkan:2004} is a graph-based approach that computes sentence importance based on the concept of eigenvector centrality in a graph representation of source sentences.

\item \textbf{TextRank}~\cite{Mihalcea:2004} is an unsupervised graph-based ranking algorithm inspired by algorithms PageRank and HITS.

\item \textbf{SumBasic}~\cite{Vanderwende:2007} is an extractive approach that assumes words occurring frequently in a document cluster have a higher chance of being included in the summary.

\item \textbf{KL-Sum}~\cite{Haghighi:2009} describes a method that greedily adds sentences to the summary so long as it decreases the KL divergence.

\item \textbf{Distraction-M3}~\cite{Chen:2016} trains the summarization model to not only attend to
to specific regions of input documents, but also distract the attention to traverse different content of the source document.

\item \textbf{Pointer-Generator}~\cite{See:2017} allows the system to not only copy words from the source text via pointing but also generate novel words through the generator. 

\item \textbf{Graph-based Attention}~\cite{Tan:2017} introduces a graph-based attention mechanism to enhance the encoder-decoder framework.
\end{itemize}

\begin{table}[t]
\setlength{\tabcolsep}{4.5pt}
\renewcommand{\arraystretch}{1.1}
\centering
\begin{small}
\begin{tabular}{|l|rrrrr|}
\hline
& \textbf{K1} & \textbf{K2} & \textbf{K3} & \textbf{K4} & \textbf{K5}\\
\hline
\hline
\# Uniq Entities & 23.7K & 37.0K & 46.1K & 50.3K & 50.3K\\
Train Acc (\%) & 46.1 & 37.2 & 34.2 & 33.6 & 34.8 \\
Valid Acc (\%) & 12.8 & 14.0 & 14.7 & 15.7 & 15.4 \\
Valid R-2 F (\%) & {11.2} & {11.1} & {11.2} & {11.1} & {10.8}\\
\hline
\hline
\# Uniq Keywds & 7.3K & 10.4K & 12.5K & 13.7K & 13.7K\\
Train Acc (\%) & 30.5 & 28.2 & 27.6 & 27.5 & 27.5 \\
Valid Acc (\%) & 19.3 & 22.5 & 22.2 & 23.0 & 21.9 \\
Valid R-2 F (\%) & {11.2} & {11.1} & {10.8} & {11.0} & {10.8}\\
\hline
\end{tabular}
\end{small}
\caption{Train/valid accuracy and R-2 F-scores when using varying numbers of QA pairs (K=1 to 5) in the reward func.}
\label{tab:results_valid}
\vspace{-0.1in}
\end{table}

In Table~\ref{tab:results_valid}, we vary the number of QA pairs used per article in the reward function ($K$=1 to 5).
The summaries are encouraged to contain comprehensive content useful for answering \emph{all} questions.
When more QA pairs are used (K1$\rightarrow$K5), we observe that the number of answer tokens has increased and almost doubled: 23.7K (K1)$\rightarrow$50.3K (K5) for entities as answers, and 7.3K$\rightarrow$13.7K for keywords.
The enlarged answer space has an impact on QA accuracies.
When using entities as answers, the training accuracy is 34.8\% (Q5) and validation is 15.4\% (Q5), and there appears to be a considerable gap between the two.
In contrast, the gap is quite small when using keywords as answers (27.5\% and 21.9\% for Q5), suggesting that using sentence root words as answers is a more viable strategy to create QA pairs.

Comparing to QA studies~\cite{Chen:2016:CNN}, we remove the constraint that requires answer entities (or keywords) to reside in the source documents.
Adding this constraint improves the QA accuracy for a standard QA system.
However, because our system does not perform QA during testing (the question-answer pairs are not available for the test set) but only generate generic summaries, we do not enforce this requirement and report no testing accuracies.
We observe that the R-2 scores only present minor changes from K1 to K5.
We conjecture that more question-answer pairs do not make the summaries contain more comprehensive content because the input and the summary are relatively short; $K$=1 yields the best results.

In Table~\ref{tab:results_output}, we present example system and reference summaries.
Our extractive summaries can be overlaid with the source documents to assist people with browsing through the documents. 
In this way the summaries stay true to the original and do not contain information that was not in the source documents.

\vspace{0.05in}
\noindent\textbf{Future work.} 
We are interested in investigating approaches that automatically group selected summary segments into clusters. 
Each cluster can capture a unique aspect of the document, and clusters of text segments can be color-highlighted.
Inspired by the recent work of Narayan et al.~\shortcite{Narayan:2018}, we are also interested in conducting the usability study to test how well the summary highlights can help users quickly answer key questions about the documents. This will provide an alternative strategy for evaluating our proposed method against both extractive and abstractive baselines.

\begin{table}
\centering
\setlength{\tabcolsep}{2pt}
\renewcommand{\arraystretch}{1.1}
\begin{scriptsize}
\begin{fontppl}
\begin{tabular}{|p{3in}|}
\hline
\textbf{Source Document}\\[0.7mm]
It was all set for a fairytale \textbf{ending for record breaking jockey AP McCoy}. In the end it was a different but familiar name who \textbf{won the Grand National on} Saturday.\\ [0.7mm]
25-1 outsider Many Clouds, who had shown little form going into the race, won by a length and a half, \textbf{ridden by jockey Leighton Aspell.} \\[0.7mm]
\textbf{Aspell won last year's Grand National} too, making him \textbf{the first jockey since the 1950s to ride back-to-back winners on different horses}. \\[0.7mm]
``It feels wonderful, I asked big questions,'' Aspell said... \\[0.7mm]
\hline
\textbf{Abstract}\\[0.7mm]
25-1 shot Many Clouds wins Grand National \\
Second win a row for jockey Leighton Aspell \\
First jockey to win two in a row on different horses since 1950s\\
\hline
\end{tabular}
\end{fontppl}
\end{scriptsize}
\caption{Example system summary and human abstract. The summary words are shown in bold in the source document.}
\label{tab:results_output}
\vspace{-0.1in}
\end{table}

\section{Conclusion}
\label{sec:conclusion}

In this paper we explore a new training paradigm for extractive summarization.
Our system converts human abstracts to a set of question-answer pairs.
We use reinforcement learning to exploit the space of extractive summaries and promote summaries that are concise, fluent, and adequate for answering questions. 
Results show that our approach is effective, surpassing state-of-the-art systems.

\section*{Acknowledgments}

We thank the anonymous reviewers for their valuable suggestions. This work is in part supported by an unrestricted gift from Bosch Research. Kristjan Arumae gratefully acknowledges a travel grant provided by the National Science Foundation.

\bibliography{summ,abs_summ,fei}
\bibliographystyle{acl_natbib}

\end{document}